\documentclass[final]{cvpr}

\usepackage{times}
\usepackage{epsfig}
\usepackage{graphicx}
\usepackage{amsmath}
\usepackage{amssymb}
\usepackage{cite}

\usepackage{graphics} % for pdf, bitmapped graphics files
\usepackage{graphicx}
\usepackage{grffile} %file name with multi-dots
\usepackage{amsmath} % assumes amsmath package installed
\usepackage{amssymb}  % assumes amsmath package installed
\usepackage{color}

\usepackage{bm}
\usepackage{url}

\usepackage[noend]{algpseudocode}
\usepackage{algorithm}
\usepackage{makecell}
\usepackage[square, comma,numbers,sort&compress]{natbib}
\usepackage{multirow}
\usepackage{enumitem}
\usepackage[table,dvipsnames]{xcolor}
\usepackage{xcolor}
\usepackage[pagebackref=true,breaklinks=true,colorlinks,bookmarks=false]{hyperref}

\hypersetup{
linkcolor=BrickRed
,citecolor=Green
,filecolor=Mulberry
,urlcolor=NavyBlue
,menucolor=BrickRed
,runcolor=Mulberry
,linkbordercolor=BrickRed
,citebordercolor=Green
,filebordercolor=Mulberry
,urlbordercolor=NavyBlue
,menubordercolor=BrickRed
,runbordercolor=Mulberry
}

\def\confYear{CVPR}

\begin{document}

%%%%%%%%% TITLE and ABSTRACT
% \title{Deep Clustering Benchmarks for CAD Models}
% \def\cvprPaperID{11112}
% \title{Cluster3D: A Dataset and Benchmark for \\ Clustering Non-Categorical 3D CAD Models
% }
% \title{Cluster3D: A Dataset and Protocol for\\Evaluating Clustering Algorithms for Non-Categorical 3D CAD Models
% }

\title{Evaluating Deep Clustering Algorithms on Non-Categorical 3D CAD Models}

% \author{
%  Siyuan Xiang$^{1}$\thanks{Equal contributions. } \and
%  Chin Tseng$^{1}$\footnotemark[1] \and
%  Congcong Wen$^{1}$ \and
%  Deshana Desai$^{1}$ \and
%  Yifeng Kou$^{1}$ \and
%  Binil Starly$^{2}$ \and
%  Daniele Panozzo$^{1}$ \and
%  {Chen Feng\thanks{The corresponding author is Chen Feng {\tt\small cfeng@nyu.edu}.} \ }
%  \\
%  \textbf{\url{https://ai4ce.github.io/SPARE3D}}
% }

\author{
 {Siyuan Xiang\thanks{Equal contribution.} \ \footnotemark[3]{}}
 \and
 {Chin Tseng\footnotemark[1]{} \ \footnotemark[3]{}}
 \and
 {Congcong Wen\footnotemark[3]{}}
 \and
 {Deshana Desai\footnotemark[3]{}}
 \and 
 {Yifeng Kou\footnotemark[3]{}}
 \and 
 {Binil Starly\footnotemark[4]{}}
 \and 
 {Daniele Panozzo\footnotemark[3]{}}
 \and
 {Chen Feng\thanks{The corresponding author is Chen Feng {\tt\small cfeng@nyu.edu}.} \
 \footnotemark[3]{}}
 \and 
 \textsuperscript{$\ddagger$}New York University
 \quad 
 \textsuperscript{$\mathsection$}Arizona State University
 \\ \url{https://cluster3d.github.io/}
}

\maketitle
% \end{document}

% \author{%
%   David S.~Hippocampus\thanks{Use footnote for providing further information
%     about author (webpage, alternative address)---\emph{not} for acknowledging
%     funding agencies.} \\
%   Department of Computer Science\\
%   Cranberry-Lemon University\\
%   Pittsburgh, PA 15213 \\
%   \texttt{hippo@cs.cranberry-lemon.edu} \\
%   % examples of more authors
%   % \And
%   % Coauthor \\
%   % Affiliation \\
%   % Address \\
%   % \texttt{email} \\
%   % \AND
%   % Coauthor \\
%   % Affiliation \\
%   % Address \\
%   % \texttt{email} \\
%   % \And
%   % Coauthor \\
%   % Affiliation \\
%   % Address \\
%   % \texttt{email} \\
%   % \And
%   % Coauthor \\
%   % Affiliation \\
%   % Address \\
%   % \texttt{email} \\
% }

% \twocolumn[{%
% \maketitle
% \vspace{-10mm}
% \begin{figure}[H]
% \begin{center}
%     \hsize=\textwidth 
%     \includegraphics[width=0.8\textwidth]{figs/fig1.png}
%     \caption{\textbf{The overview of Cluster3D} via a subset of clustering results from a baseline (DeepCluster~\cite{caron2018deep}) demonstrating the challenges of classification-based labeling in our task, due to many non-standard mechanical components (green). Each section shows some random CAD models in the same cluster. A red section shows a cluster with annotation violations highlighted at the red objects.}
%     \label{fig:overview}
% \end{center} 
% \end{figure}
% }]
% \vspace{-1cm}

% \maketitle

% \maketitle
\thispagestyle{empty}

%%%%%%%%% BODY TEXT
\begin{abstract}
% benchmark paper, proposing new evaluation metric for clustering on 3D CAD models
% Clustering is an approach to partition data into a certain number of clusters, and the similar data are inclined to be clustered into the same cluster. This property can help learn the data's nature structure and distribution if we do not have prior knowledge about the data. Inspired by some existing traditional clustering methods or deep clustering methods for 2D images, we adapt these methods to a non-categorical 3D CAD model dataset. However, without knowing the classes, it becomes a problem to evaluate these clustering methods. In this work, we provide a similarity matrix which can be used to evaluate the clustering performances. The similarity matrix is annotated by experts in mechanical domain, and we use their domain experience to judge the similarity between CAD models. 
% \begin{figure*}[h]
% \vspace{-3mm}
%     \centering
%     \includegraphics[width=0.8\textwidth]{figs/fig1.png}
%     \caption{\textbf{The overview of Cluster3D} via a subset of clustering results from a baseline (DeepCluster~\cite{caron2018deep}), demonstrating the challenges of classification-based labeling in our task, due to many non-standard mechanical components (green). Each section shows some random CAD models in the same cluster. A red section shows a cluster with annotation violations highlighted at the red objects.}
%     \label{fig:overview}
%     \vspace{-3mm}
% \end{figure*}

% We introduce the first large-scale non-categorical 3D CAD dataset to study 3D deep clustering and tackle its evaluation challenge. 
We introduce the first work on benchmarking and evaluating deep clustering algorithms on large-scale non-categorical 3D CAD models.
We first propose a workflow to allow expert mechanical engineers to efficiently annotate $252,648$ carefully sampled pairwise CAD model similarities, from a subset of the ABC dataset with $22,968$ shapes. 
Using seven baseline deep clustering methods, we then investigate the fundamental challenges of evaluating clustering methods for non-categorical data. Based on these challenges, we propose a novel and viable ensemble-based clustering comparison approach.
This work is the first to directly target the underexplored area of deep clustering algorithms for 3D shapes, and we believe it will be an important building block to analyze and utilize the massive 3D shape collections that are starting to appear in deep geometric computing. 

% \keywords{Deep clustering, unsupervised learning, evaluation protocol}

\vspace{-3mm}

\end{abstract}
\section{Introduction}\label{sec:intro}

\begin{figure*}[h]
\vspace{-3mm}
    \centering
    \includegraphics[width=0.9\textwidth]{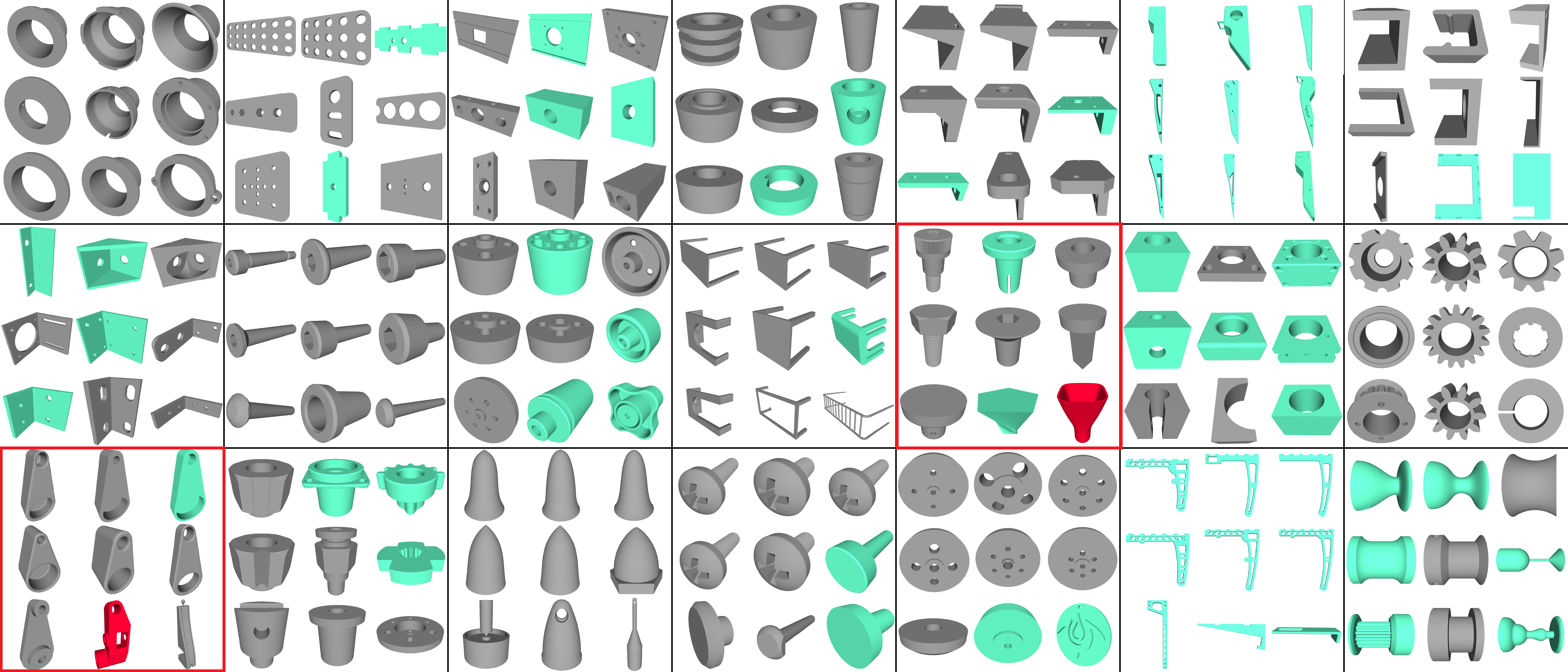}
    \caption{\textbf{The overview of our work} via a subset of clustering results from a baseline (DeepCluster~\cite{caron2018deep}). It demonstrates the challenges of classification-based labeling in our task, due to many non-standard mechanical components (green). Each section shows some random CAD models in the same cluster. A red section shows a cluster with annotation violations highlighted at the red objects.}
    \label{fig:overview}
    \vspace{-3mm}
\end{figure*}

% Think:/

% 1. what is your figure 1?

% 2. what is your overview figure (figure 2)?

% Tips:

% 1. never label your figures/tables as ``\textbackslash label\{fig1\}'', label it by a short-name such as ``\textbackslash label\{fig:teaser\}'', or ``\textbackslash label\{tab:main\_results\}''.

% 2. Quote-marks in \LaTeX should be: \`{}\`{} for left quote-mark and two \'{}\'{} for right quote-mark.

% 3. Try to reduce orphan words (\url{https://www.pinterest.ph/pin/158963061822034215/}).
Clustering and classification are two ways to recognize repeated patterns in large-scale 3D model datasets, which could serve downstream applications like 3D data management, compression, search, and exploration~\cite{4267943, 6200340, 8648155, 10.1145/3343031.3351061}. Currently, the community is focusing more on categorical dataset~\cite{chang2015shapenet, wu20153d} for 3D object classification tasks~\cite{qi2017pointnet, socher2012convolutional, qi2016volumetric}.
However, sometimes the data distribution, especially in large-scale CAD datasets, might be very complex, and class labels are very difficult to assign, making it hard to train and evaluate shape classification.
ABC dataset~\cite{koch2019abc} is an example of such type of dataset, which is composed of 1 million 3D CAD models manually modeled by hobbyists and experts alike. \textit{One may easily understand the difficulty of the classification task by trying to assign names to each group of objects in Figure~\ref{fig:overview}}, and more challenge details are in the Section~\ref{sec:related} and supplementary. Therefore, we have to explore clustering-based analysis of such datasets.

To the best of our knowledge, we are the first to explore the deep clustering algorithms on such a non-categorical 3D model dataset. We first establish a benchmark, adopting classic or state-of-the-art deep clustering algorithms on the dataset. More importantly, we propose a strategy for evaluating these methods. Previously, the external evaluation principles for most of the clustering algorithms in vision and learning community are based on class labels~\cite{caron2018deep, zhan2020online}, which oversimplifies the inter-cluster and intra-cluster object similarity and does not apply to our case. Differently, our evaluation strategy directly examines the pairwise similarity relationships between every two CAD models. 
For this evaluation system, we propose a scalable and effective pairwise similarity annotation workflow and implement a graphical user interface for efficient human annotations. Based on the pairwise similarity, we propose a novel ensemble-based clustering evaluation protocol to rank the clustering algorithms.

For creating the deep clustering benchmark on non-categorical 3D models, we select a subset in ABC dataset~\cite{koch2019abc}, adapting seven unsupervised deep clustering methods to the subset.
These clustering methods can be classified into two types: (1) two-stage clustering and (2) end-to-end deep clustering. For the two-stage clustering approach, we first perform deep representation learning on 3D mechanical components to extract high dimensional features, using pre-trained neural networks. Then we apply classic clustering algorithms, like KMeans~\cite{macqueen1967some}, to group these learned features into different clusters. For end-to-end deep clustering methods, we combine representation learning, dimensionality reduction, and clustering in an end-to-end framework. 

For the evaluation protocol, instead of using class labels, we ask expert annotators (graduate students in mechanical engineering) to label pairwise 3D shape similarity relationships between each pairs of 3D models as the external knowledge for evaluation. But annotating all the pairs is obviously intractable: a rough estimate is that a single annotator working 24 hours a day would need 16 years to annotate the $500$ millions of edges in our dataset!
As a compromise, we propose a workflow to carefully select a small subset of non-trivial pairs to obtain useful annotations. We developed a web-based user interface using this workflow and annotated $252,648$ selected pairwise similarities on $22,968$ ABC shapes.

However, because of the relatively sparse annotation compared to annotating all the pairs, we find that bias is inevitable so we cannot rely only on these human annotations for a fair evaluation of baselines.
Trying to tackle \textit{this dilemma, where full annotation is intractable and sparse annotation is unfair}, we propose a novel clustering evaluation protocol based on the ensemble principle~\cite{zhang2012ensemble}.
% \DP{cite for ensemble?}.

Instead of trying to find an ``absolute'' ranking of baselines by comparing them to an unobservable ground truth, \textit{our evaluation protocol is ``relative, dynamic, and democratic''} by using the ensemble of different similarity predictions and human annotations as an approximation of that ground truth. Our justification for this evaluation protocol is based on the general consensus of the reduced variance and improved robustness/accuracy of ensemble-based system~\cite{zhang2012ensemble}, and also our various experimental evidences. 

We believe that our evaluation strategy is useful to provide an objective metric on clustering tasks on non-categorical dataset. \textit{Because of the uniqueness of our protocol, we plan to continue collecting data and clustering methods, periodically updating the benchmark}.
The human annotations and benchmark results, the annotation software, all baseline implementations, and the evaluation scripts will be publicly released as open source using the MIT license.

In summary, our contributions are the following:
\begin{itemize}
    \item To the best of our knowledge, it is the first work focusing on deep clustering for non-categorical 3D shapes, which could stimulate a new direction for 3D deep clustering.

    \item We propose a scalable and effective pairwise similarity annotation workflow, implemented in a graphical user interface, to allow experts to efficiently label a large number of non-trivial 3D object pairwise similarity (for a total of $252,648$ annotation pairs per annotator). 
    
    \item We adapt 7 deep clustering algorithms for 3D CAD models, creating the first 3D deep clustering benchmark.
    
    \item We propose a novel ensemble-based clustering evaluation protocol for non-categorical data, with experimental justifications using 7 baseline clustering methods and several internal evaluation indices.

\end{itemize}

\section{Related Work}\label{sec:related}

% Different way of citing a paper:

% \cite{Alpher02}.
% \citet{yang2018foldingnet}.

% \citeauthor{yang2018foldingnet}.

% \citeyear{yang2018foldingnet}.

% Dataset | ME? | #Models | #Tasks | #Dependent | 

We cover the related works more closely related to our main contributions: (1) clustering algorithms and corresponding evaluation metrics, (2) large datasets of 3D models, and (3) approaches for annotating 3D datasets.

\section{Annotation Creation}\label{sec:annotation}

% We introduce a system to evaluate deep clustering on large-scale 3D CAD models based on an annotated similarity matrix. 
%We first discuss the annotation process to create the  similarity matrix (Section \ref{}), and then introduce seven clustering methods to which can serve as a benchmark results. Next, we propose our evaluation metrics using the annotated similarity matrix.

\subsection{Human annotation}
We view the selected subset from ABC dataset as an undirected complete graph $\mathcal{G}(\mathcal{V}, \mathcal{E})$. The node set $\mathcal{V}$ contains all the 3D CAD models in the subset, each one is a node $v\in\mathcal{V}$. We denote the cardinality of the set as $|\mathcal{V}|$. An edge $e_{i,j}\in\{+1,-1,0\}$ in the edge set $\mathcal{E}$ stores the similarity annotation of two 3D CAD models $v_i$ and $v_j$. The edge labels $+1, -1$, and $0$ respectively indicate similar, dissimilar, and unknown relationship between two nodes. The edge set can come from human annotation or from a clustering algorithm. Next, we introduce how we create and annotate the subset, and then discuss several important design considerations.

\subsubsection{Annotation creation workflow}\label{sec:workflow}

\begin{figure*}[t]
    \centering
    \includegraphics[width=0.9\textwidth]{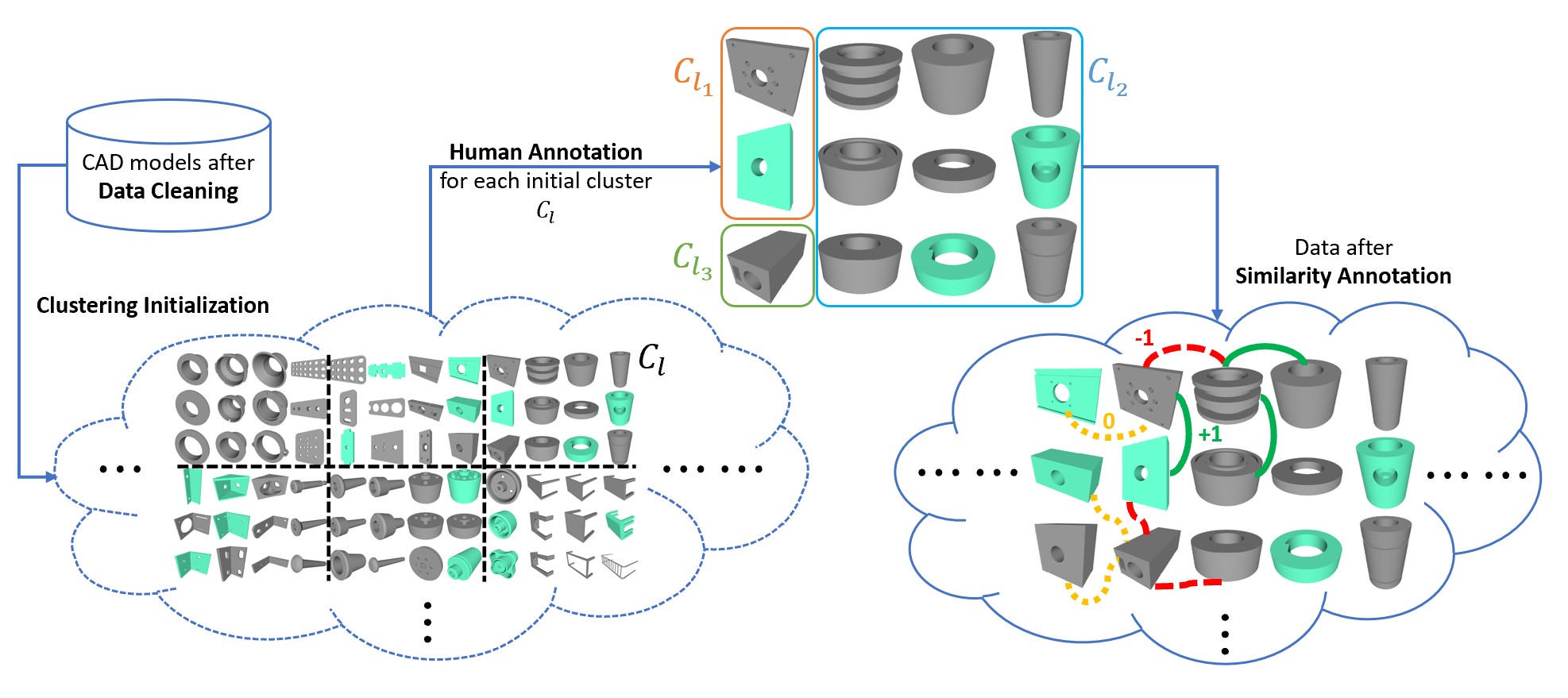}
    \caption{Human annotation workflow.}
    % \vspace{-5mm}
    \label{fig:workflow}
\end{figure*}

We use the workflow illustrated in Figure~\ref{fig:workflow} with the following major steps:

\textbf{Step 1: Data Cleaning.} We use the first four chunks of the ABC dataset~\cite{koch2019abc}, and filter out all blank files and all files containing assemblies instead of single components, obtaining a subset of $22,968$ CAD models.

\textbf{Step 2: Similarity Annotation.}
While ideally we would like to manually annotate each edge, this is not practical. We thus introduce a method to carefully sample a subset of edges that will be manually annotated. 

\textit{Step 2.1: Cluster Initialization.}
We group all the CAD models into a set of initial clusters $\mathcal{C_I}\equiv \{\mathcal{C}_k|\cup_{\forall k} \mathcal{C}_k=\mathcal{V}; \mathcal{C}_k\cap \mathcal{C}_l=\varnothing, \forall k\neq l; \text{and } |\mathcal{C}_k|\leq T, \forall k\}$, each containing no more than $T=12$ CAD models, using a clustering method detailed in Section~\ref{sec:annotation-decisions}. The number $12$ is chosen to facilitate a comfortable and useful labeling by humans in the graphical interface, because they can easily see the details of each model and still be able to compare each other efficiently at a glimpse. We then automatically assigned label 0 (meaning unknown) to all the edges across different clusters, i.e., $e_{i,j}=0 \iff v_i \in \mathcal{C}_k, v_j \in \mathcal{C}_l, \text{and } k \neq l$. 

\textit{Step 2.2: Human Annotation.}
We only manually annotate edges residing inside the same initial cluster, i.e., 
$e_{i,j}\neq 0 \iff \exists \mathcal{C}_k\in\mathcal{C_I}, v_i \in \mathcal{C}_k, v_j \in \mathcal{C}_k$.
% $e_{i,j}\neq 0 \iff \exists k\in[1,K_I], v_i \in \mathcal{C}_k, v_j \in \mathcal{C}_k$, 
A number of $A=4$ mechanical engineers served as our experts to provide their CAD model similarity annotations independently. For each initial cluster, e.g., $\mathcal{C}_k$, an annotator has to either \textit{confirm} that CAD models inside $\mathcal{C}_k$ are all similar to each other, or further \textit{divide} the cluster into smaller clusters until such confirmations can be made for each smaller cluster. The \textit{confirmation} of a cluster $\mathcal{C}_l$ assigns all internal edges inside this cluster with the \textit{positive} label $+1$, i.e., $e_{i,j}=+1 \iff v_i \in \mathcal{C}_l, v_j \in \mathcal{C}_l$. The \textit{division} of a cluster $\mathcal{C}_l$ into smaller clusters $\{\mathcal{C}_{l_t}|\cup_{\forall t}~\mathcal{C}_{l_t} = \mathcal{C}_l; \mathcal{C}_{l_s}\cap \mathcal{C}_{l_t}=\varnothing, \forall s\neq t\}$ assigns all edges across those small clusters with the \textit{negative} label $-1$, i.e., $e_{i,j}=-1 \iff v_i \in \mathcal{C}_{l_t}, v_j \in \mathcal{C}_{l_s}, s\neq t$. We record each annotator independently, i.e., the $a$-th annotator's annotation forms an edge set $\mathcal{E}^a$ over the same nodes. The instruction information is in the supplementary.

\subsection{Design Decisions on the Annotation Workflow}\label{sec:annotation-decisions}

Although it has been widely used in geometry processing and machine learning, the concept of similarity can be vague and ambiguous when applied to 3D CAD models in Cluster3D.

\textbf{Why manually annotate similarity?}
To determine whether two 3D models are similar or not algorithmically, 
there are two main criteria: geometric distribution similarity~\cite{shum19963d, hilaga2001topology, funkhouser2003search, osada2002shape} and visual similarity~\cite{chen2003visual}. Yet for human beings, the mechanism to determine the similarity between two CAD models is also based on unconscious background knowledge~\cite{yang2014object, de2008perceived}, which might be different from the similarity judgment encoded in existing algorithms~\cite{yang2014object, cutzu1998representation}. Therefore, 
% even with mathematically defined 3D object similarity metrics, 
acquiring large-scale human annotations for pairwise CAD model similarity is still important to capture experts' underlying background knowledge. In our practice, annotators are only given 3D CAD models, without extra information like material or size. Therefore, in most cases human annotation is determined by the geometric information. 

\textbf{Reasons for cluster initialization}.
It is impractical for human experts to annotate the similarity relationships between all pairs of CAD models, as their number is quadratic with respect to the number of models considered. It would take 16 years of annotation time, assuming a 1 second time to annotate each edge. 

Therefore, we have to commit to annotate only a small subset of edges. The most obvious and unbiased approach is to random sample from all the edges. This is however not an option for our case, as the sampling would be too unbalanced, as most of the edges indicate dissimilarity between objects. 
% due to the non-categorical nature \DP{again lets discuss this non categorical term} of our data. 
We needed a strategy that would give us a more or less even split between similar and dissimilar edges so that we could use our expert time meaningfully and evaluate clustering methods in a balanced way.

After experimenting with different approaches, we found a method that leads to a reasonable 1 to 1 ratio of similar and dissimilar edges in our dataset: we use a clustering algorithm to over segment the dataset in small clusters (12 objects/cluster), and then ask users to annotate all edges within each cluster. Overall, this approach allowed us to get a good distribution of edge labels while annotating about only 0.5\% of the entries in our similarity matrix, making the annotation problem tractable with our budget and resources.

\textbf{Cluster initialization and evaluation bias}.
A natural question is how to use this extremely sparse annotation to evaluate baseline clustering methods, because the different choices of cluster initialization would lead to a different subset of annotated edges. Will this inevitably introducing bias if we directly use the annotations for evaluation? Or will we obtain consistent evaluation even if the cluster initialization is different? \textit{To answer this question in our experiment section, we use two different methods for cluster initialization}: the MVCNN~\cite{su2015multi}-based method and the AtlasNet~\cite{groueix2018papier}-based method, which leads to different subsets of edges for annotation.

% We use two different methods for cluster initialization: the MVCNN~\cite{su2015multi}-based method and the AtlasNet~\cite{groueix2018papier}-based method. This leads to different annotated subsets, allowing us to verify: (1) whether clustering evaluation using only a small subset of human annotations will introduce bias, and (2) if a method can consistently perform well on different human annotated subsets.

\textit{For the MVCNN-based method}, we first generate $12$ images for all $22,968$ CAD models in our dataset, following the original settings in~\cite{su2015multi}. We use all these $12 \times 22,968$ images to train a convolutional auto-encoder network. Then the trained encoder is used to extract features for all these images. For each CAD model, we concatenate the twelve latent vectors from its corresponding $12$ images to represent its features. 
\textit{For the AtlasNet-based method}, we follow the original auto-encoder architecture to reconstruct 3D point cloud for each input 3D CAD object, and then predict the features for each CAD model using the encoder of the trained model.

Finally, for both methods, $22,968$ features representing all shapes in Cluster3D are clustered by KMeans algorithm.
With the number of clusters $K$ in KMeans set to $2,000$, we obtain two thousand initial clusters for the human annotators. We continue to split the clusters with more than 12 models in the class using KMeans, until the contained number of models is no greater than $12$. These clustered CAD models are then loaded into our annotation interface.

% \DP{This is abrupt, I was expecting here a discussion of the bias or the lack thereof.}

\textbf{Annotation interface.}
We developed a web-based annotation application. 
The interface shows CAD models of one cluster at a time. It shows the 12 CAD models with checkbox, and initially all 12 checkbox are set to be checked. The annotators manually unmark the models which are considered dissimilar from the others, effectively annotating all edges linking the 12 models in the cluster. After confirmation, a new set of 12 models is shown. We will show our web interface in the supplementary.

\textbf{Data statistics.}
The selected CAD model subset has $22,968$ CAD models; therefore, there are $22,968 \times 22,967 / 2$ similar or dissimilar edges in total between every two CAD models. We have four human experts to annotate two different edge subsets due to different cluster initialization, and each initial cluster contain $252,648$ edges to be labeled. This leads to eight different human annotation edge sets $\mathcal{E}^a$. The statistics of average annotation consistency between different human experts is above $70 \%$; therefore, we believe it is reasonable to use these human annotations, especially after using majority voting strategy (details in Section~\ref{sec:ensemble}). 
More statistics for annotation consistency between different human experts, \textit{positive} and \textit{negative} labels for each expert can be found in the supplementary.
% Among them, $252,648$ edges are labeled by three human experts respectively. For the first annotator, $155,960$ edges are labeled as $1$, representing these CAD model pairs are similar, and $119,656$ are labeled as $-1$, meaning these pairs are dissimilar. For the second and third annotator, they have labeled $130\,442$, $145\,174$ similar edges, and $205\,582$, $70\,034$ dissimilar edges respectively. We also check the consistency of the three annotators' labeling. If consistency means the three annotations for a specific edge are all the same, then the total number of consistent label is $172,554$, occupying $62.6 \%$ of the labeled edges. More details can be found in the supplementary that show a decent annotation consistency when defining it as the majority vote.

% \begin{figure}[htp]
%     \centering
%     \includegraphics[width=0.5\textwidth]{figs/edge.png}
%     \caption{\textbf{Annotation Edge Counts}.}
%     \label{annotator-edge}
% \end{figure}

% \begin{figure}[htp]
%     \centering
%     \includegraphics[width=0.5\textwidth]{figs/consistency.png}
%     \caption{\textbf{Annotation Consistency}.}
%     \label{annotator-consistency}
% \end{figure}

\section{Benchmark}\label{sec:benchmark}

% Tips:

% 1. no need to number every equation. Only number the important ones and the ones that you need to refer to later.

% 2. Use ``\textbackslash eqref'' to refer to an equation. ``\textbackslash ref'' is for figures/tables.

\subsection{Baseline methods}
Based on the availability and usability of source codes, we select and adapt seven baseline methods to establish a benchmark for clustering algorithms. We divide these baseline methods into two types: (1) two-stage clustering, and (2) end-to-end deep clustering. Two-stage clustering methods use a deep neural network to extract features for all CAD models, then apply a  traditional clustering algorithm, such as KMeans. End-to-end deep clustering baseline methods integrate feature extraction and clustering in one framework: during the training process both network loss and clustering loss are minimized. Note that all these methods are considered as partitional clustering~\cite{celebi2014partitional}, i.e. one CAD model will only fall into one cluster.

Since some of the baseline methods are designed for 2D images, we adapt them for 3D CAD models. We use either point cloud or multi-view images as the input representation, and select suitable deep neural networks. A detailed description of all network architectures can be found in the supplementary.

% \textit{AtlasNet~\cite{groueix2018papier}:} To compute cluster of 3D point clouds using Atlasnet, we follow the original auto-encoder architecture to reconstruct 3D point cloud for each input 3D CAD object, and then predict latent vectors based on the encoder of the trained model. The CAD objects are clustered by using KMeans on the obtained latent features.

\textbf{Two-stage clustering.} 
Also used for cluster initialization, \textit{MVCNN~\cite{su2015multi}} and \textit{AtlasNet~\cite{groueix2018papier}} have been described in Section \ref{sec:annotation}. We add one more  baseline in this group, \textit{BYOL~\cite{grill2020bootstrap}:} BYOL is proposed to compute self-supervised image representation learning. We replace the image encoder (ResNet) with a point cloud encoder (PointNet) to learn a representation of a 3D CAD shape. We use three types of data augmentation on point cloud: random dropout, random scaling, and random shift.
We then apply the KMeans algorithm on the learned latent representation to cluster CAD objects.

\textbf{End-to-end deep clustering.} Four methods are selected in this group.

\textit{DEC~\cite{xie2016unsupervised}:} To adapt the DEC algorithm, we initialize DEC with the AtlasNet architecture to auto-encode 3D point clouds as the input data. The deep auto-encoder is trained to minimize Chamfer loss and learns representations of the 3D shapes. We then follow the DEC algorithm by discarding the decoder layers and use the encoder layers as the initial mapping between the data and feature space. This is followed by joint optimization of the cluster centers and encoder parameters using SGD with momentum.

\textit{DeepCluster~\cite{caron2018deep}:} We replace the convolution networks trained by the DeepClustering algorithm to use PointNet instead for encoding the point cloud data to predict cluster assignments. The algorithm is followed by alternating between clustering of the point cloud feature descriptors using K-Means and training the PointNet network using the multinomial logistic loss function.

\textit{IIC~\cite{ji2019invariant}:} Instead of the original IIC method for unsupervised image semantic task, we first randomly transform a CAD model to a pair of point clouds, and use PointNet as encoder to maximize mutual information between the class assignments of each pair. The trained model directly outputs class labels for each 3D CAD model.

\textit{SCAN~\cite{van2020scan}:} We adjust the pretext stage: Instead of using noise contrastive estimation (NCE) to determine the nearest neighbors, we use the auto-encoder of AtlasNet we have trained to output the feature vectors to generate the nearest neighbors set.

\subsection{Evaluation Metrics}

As discussed in Section~\ref{sec:related}, external and internal evaluation indices are used for evaluating clustering results.

\textbf{External evaluation indices.}
% With the annotated similarity matrix $G$, we have the prior knowledge to determine if two CAD objects are similar, dissimilar, or unknown, as shown in Figure~\ref{similarity_matrix}. 
We evaluate the clustering results of the baseline methods using two external evaluation indices:  \textit{edge accuracy} and \textit{balanced accuracy}. \textit{Edge accuracy} is the binary similarity classification accuracy for each edge compared to a ground truth similarity matrix. For the clustering results obtained from the baseline methods, the CAD models grouped into the same cluster are considered as pairwise similar, thus predicting the edges between these models as $+1$. The predicted edges between models in different clusters are considered as $-1$. We calculate the average classification accuracy for all the edges as \textit{edge accuracy}.
However, in practice, most of the CAD models are dissimilar to each other, which means most of the edges have ground truth labels as $-1$. Therefore, the dataset is heavily imbalanced. To better evaluate the baseline methods' performance, we use balanced accuracy~\cite{brodersen2010balanced}, which suffer less from the imbalance. More calculation details are available in the supplementary.

\begin{figure*}[htp]
    \centering
    \includegraphics[width=0.9\textwidth]{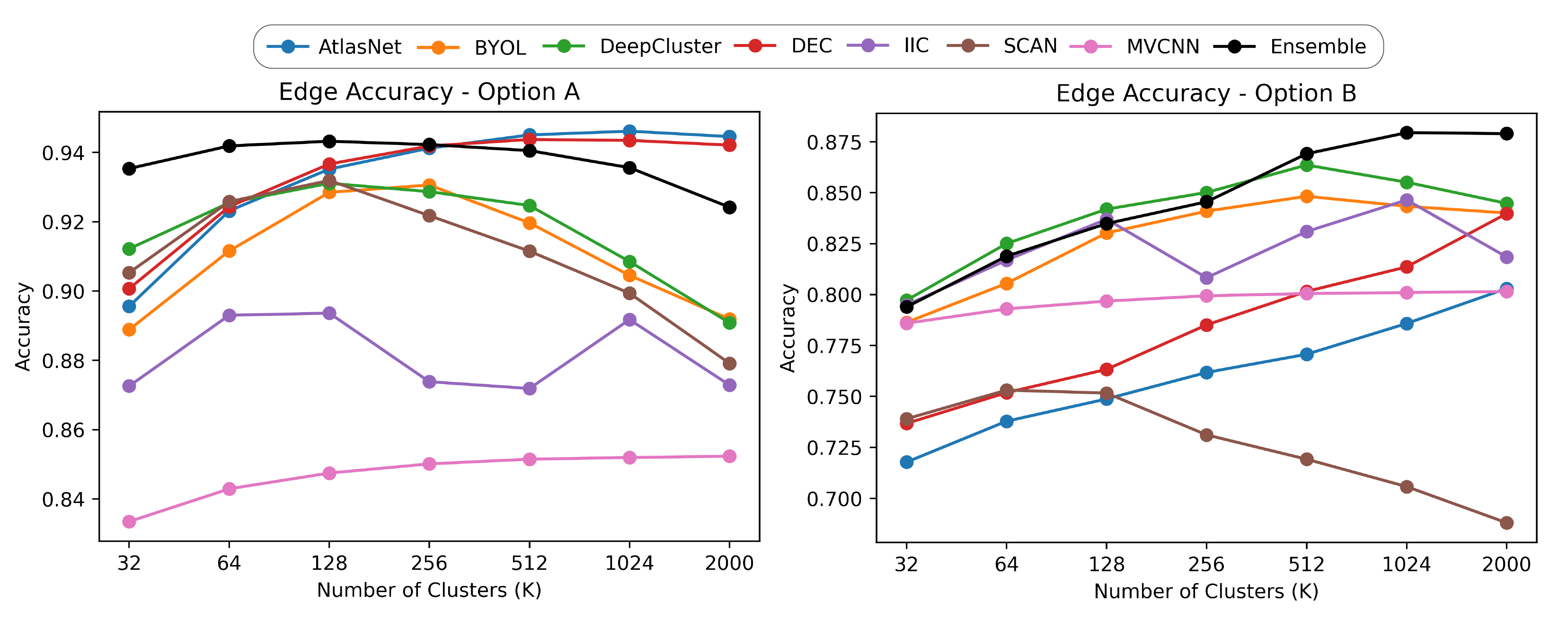}
    \caption{\textbf{Evaluation bias when using \emph{Human} alone}, based on different cluster initialization from MVCNN (option-A) and AtlasNet (option-B).}
    \label{ensemble-evaluation}
    \vspace{-5mm}
\end{figure*}

\textbf{Internal evaluation indices.}
%Internal validation indices can be used to evaluate the clustering results. 
%
We opt for the silhouette coefficient~\cite{rousseeuw1987silhouettes} method, which is widely used in clustering problems.
Based on its definition, we need to determine the distance between every two objects. In our dataset, the objects are 3D CAD models which can be represented as point cloud or voxel grids. Therefore, we choose to use the Chamfer distance~\cite{fan2017point} and the Jaccard distance~\cite{kosub2019note} as two distances between CAD models.

\subsection{Ensemble-based Evaluation}\label{sec:ensemble}

As discussed, when we use external evaluation indices, a ground truth similarity matrix is needed. Because the full ground truth is intractable to be annotated, one naive way is to directly use the sparsely annotated similarity matrix from Section~\ref{sec:workflow}, which could lead to biased evaluation. Another way is to ensemble multiple similarity matrices as a proxy of the full ground truth matrix.

The ensemble similarity matrix (denoted \emph{Ensemble}) is generated by combining $n$ different similarity matrices using a majority voting strategy.
Similarly as the human annotation result matrix, the \emph{Ensemble} matrix can be viewed as an edge set over the graph.
% $\mathcal{G}(\mathcal{V}, \mathcal{E})$, with node set $\mathcal{V}$. 
However, unlike the sparse human annotation matrix where many entries are zero, any clustering method predicts a dense matrix where each edge is known, and the edge labels either indicate similar, or dissimilar, as $e_{i,j}\in\{+1,-1\}$. Next, we will describe how we create the \emph{Ensemble} matrix, and why we think it is reasonable to use it.

\textbf{Individual similarity matrix.}
In this paper, each individual similarity matrix is independently created by the clustering result of a selected clustering method. For example, a clustering method can group all the CAD models into non-overlapping $K$ groups, as $\{\mathcal{C}_k|\cup_{\forall k} \mathcal{C}_k=\mathcal{V}, \mathcal{C}_k\cap, \mathcal{C}_l=\varnothing~ \text{and }  \forall k\neq l, \forall k\}$. Within each group $\mathcal{C}_k$, all the CAD models are considered as similar to each other, thus, $e_{i,j}=+1 \iff v_i \in \mathcal{C}_k, v_j \in \mathcal{C}_k$. Meanwhile, the CAD models in different groups are considered as dissimilar to each other, thus the edge labels are \textit{negative}, $e_{i,j}=-1 \iff v_i \in \mathcal{C}_k, v_j \in \mathcal{C}_l, k \neq l$.

%%% CFeng: moved from above, to be merged.
% \textbf{Conflicts in annotations.} 
% We use one single annotated similarity matrix as the final outcome of our annotation procedure. In case of conflicts between different annotators, the majority wins. Note that for the final evaluation we also consider the individual similarity matrices of the different annotators. 

\textbf{Ensemble by majority voting.}
After obtaining $N$ different $|\mathcal{V}| \times |\mathcal{V}|$ similarity matrices using various baseline clustering methods, we use the majority voting strategy to ensemble all the label decisions for each edge, leading to the \emph{Ensemble} matrix.
For each edge $e_{i,j}$, there are $N$ labels from $N$ similarity matrices. We define the final label of $e_{i,j}$ as the $+1$ if the number of \textit{positive} labels of $e_{i,j}$ $\geq \lceil \frac{N+1}{2} \rceil$, otherwise $-1$.

% We further consider the influence of human annotations to the \emph{Ensemble} matrix. 
We further consider to use human annotations for the evaluation protocol.
As mentioned, in our study, we have eight different annotations from four annotators and two different edge subsets. We create the human ensemble matrix (denoted \emph{Human}) by the same majority voting strategy. Note that if for an edge  $e_{i,j}$, four annotators annotate it as $+1$, and the remaining four annotators annotate it as $-1$, we will consider this edge as unknown with a label $0$.
Finally, we calculate the balanced accuracy using \emph{Ensemble} and \emph{Human} separately. We also average the two accuracy values, and denote it as \emph{EnsembleHuman}.
% Finally, we overwrite the entries of the \emph{Ensemble} matrix corresponding to the known edges in the human annotation ensemble matrix, producing the final ensemble similarity matrix as the proxy of the ground truth. We name it the \emph{EnsembleHuman} matrix.

\textbf{Justification for the ensemble-based evaluation}.
% In the previous section, we show that it is almost impossible to manually labels in our Cluster3D dataset, with more than $20,000$ CAD models. Also, bias will be inevitable introduced if we choose to the annotate a select subset, and we will show the bias in Experiment~\ref{sec:exp}. 
The ensemble by majority voting trusts the most frequently predicted edge labels among all methods/annotations. Given that the full ground truth cannot be obtained with reasonable human investment, such a proxy is a reasonable approximation for two reasons: (1) the ensemble is a commonly accepted strategy to reduce variance, enhance robustness, and therefore, improve accuracy~\cite{zhang2012ensemble} (the ensemble results are often empirically closer to ground truth) and (2) our experimental results in Figure~\ref{ensemble-evaluation} also support this choice, as we will  detail in Section~\ref{sec:exp}.

% when we evaluate the ensemble similarity matrix's performance on the two annotated human subset, and the results show the ensemble similarity matrix outperform all the single baseline methods (see section~\ref{sec:exp}). 
% Based on these two reasons, we believe it is reasonable to use ensemble similarity matrix to evaluate the baseline method's performance, since it is likely to be a good approximation of the ground truth of a dense similarity matrix.

\section{Experiments and Discussions}\label{sec:exp}

\textbf{Experiment settings.}
All the baseline methods are implemented using PyTorch~\cite{NEURIPS2019_9015} and run on an NVIDIA GeForce GTX 1080 Ti GPU. For hyperparameter settings, we tune learning rate and batch size for each baseline method. The learning rates for MVCNN-based method, AtlasNet-based method, BYOL-based method, DEC, DeepCluster, IIC, and SCAN are 0.0001, 0.001, 0.0003, 0.00001, 0.05, 0.0003, and 0.0001 respectively. The batch sizes for these methods are 60, 11, 10, 128, 50, 10, and 96 respectively. 
% \DP{why are they all different? are these the dafault parameters of each method? or did you tune them individually?}

Since we do not know a proper number of clusters in ABC, we select seven different number of clusters ($K$), from $32$ to $2,000$, following exponential growth, and we evaluate all methods on all these cases independently.

\textbf{Bias of cluster initialization.}
One question was raised in Section~\ref{sec:annotation-decisions} regarding the potential evaluation bias of using different cluster initialization methods to select the edge subsets for manual annotation. To evaluate that, we compare the similarity matrices of the human annotators initialized by the MVCNN-based method and the AtlasNet-based method to the similarity matrices obtained with various baseline methods in Figure~\ref{ensemble-evaluation}. \textit{The resulting baseline rankings are inconsistent between the two initialization methods, indicating that the choice of initial clustering method indeed introduces an evaluation bias}. This bias is due to the small number of annotated edges, and thus it is not fair to directly use the human annotated sparse similarity matrices to compare different clustering algorithms. 
% One way of removing this bias would be to annotate more edges, but this is not tractable with our limited resources.

\textbf{Empirical justification of the ensemble method.}
% We propose to use an ensemble method constructed using all our baselines as a way to reduce the bias. \DP{Explain what the ensemble method is.} 
Figure~\ref{ensemble-evaluation} also shows that ensemble method has the highest accuracy (on average over the different $K$) than all baselines. Furthermore, there are only about $6\%$ of human annotations in conflict with the result of the ensemble method, showing that it has a high degree of agreement with the human annotations. 

% We introduce two additional methods: \emph{Ensemble} which is an ensamble methods built using the 7 baseline methods, and \emph{EnsembleHuman} which additionally uses the human annotated similarities \DP{how?}. We will use these two methods as reference to evaluate the performance of the others.

\begin{figure*}[htp]
    \centering
    \includegraphics[width=0.9\textwidth]{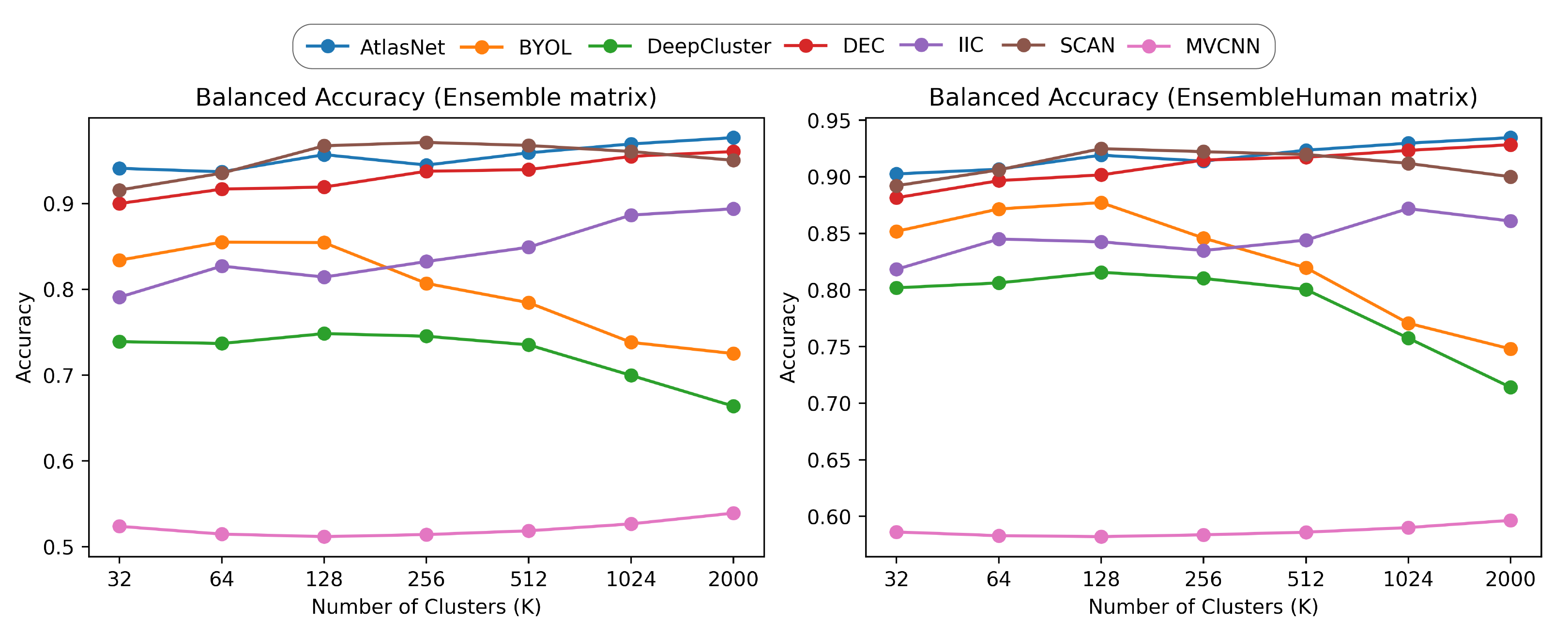}
    \caption{\textbf{Consistent ensemble-based external evaluation indices}. Rank of balanced accuracy for both \emph{Ensemble} and \emph{EnsembleHuman}: AtlasNet $>$ SCAN $>$ DEC $>$ IIC $>$ BYOL $>$ DeepCluster $>$ MVCNN.}
    \label{benchmark_results_external}
    % \vspace{-5mm}
\end{figure*}

\begin{figure*}[htp]
    \centering
    \includegraphics[width=0.9\textwidth]{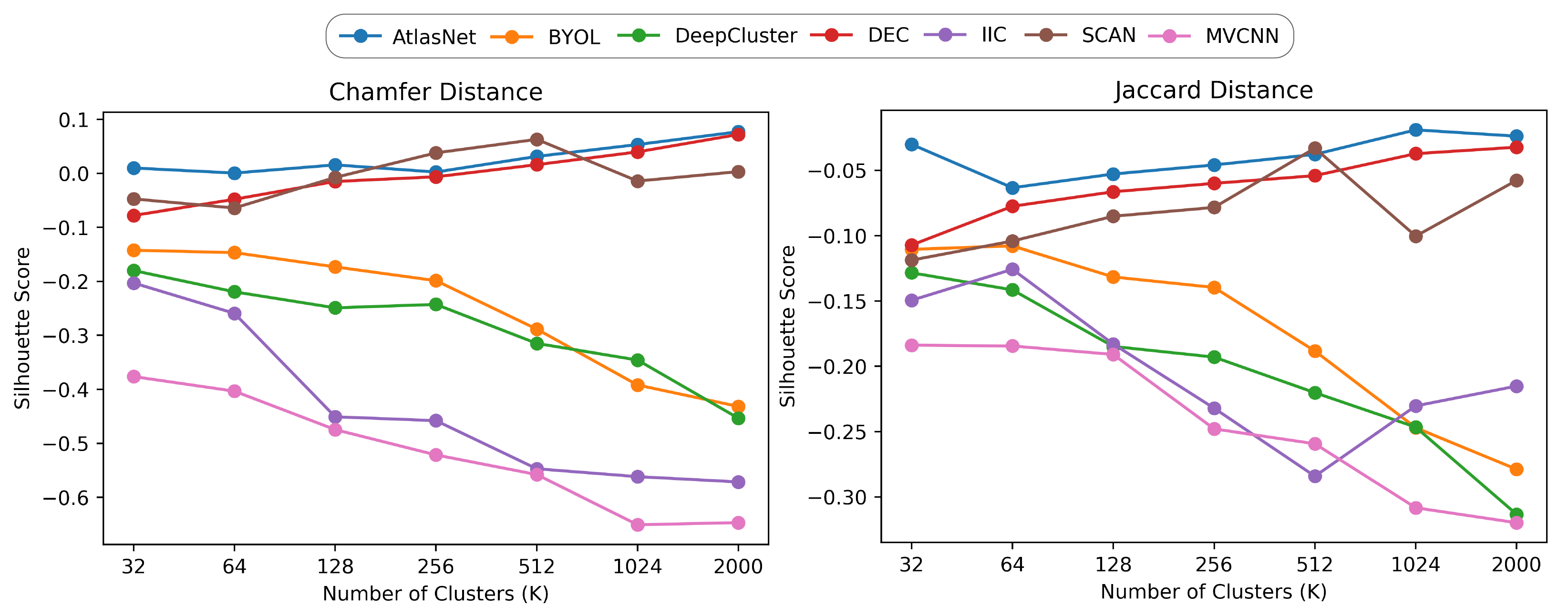}
    \caption{\textbf{Benchmark results of internal evaluation indices}. Chamfer ranking: AtlasNet $>$ SCAN $>$ DEC $>$ BYOL $>$ DeepCluster $>$ IIC $>$ MVCNN; Jaccard ranking: AtlasNet $>$ DEC $>$ SCAN $>$ BYOL $>$ IIC $>$ DeepCluster $>$ MVCNN.}
    \label{benchmark_results_internal}
\end{figure*}

\textbf{Ensemble-based external evaluation indices.}
%\siyuan{\textit{Human annotation does not have significant influence on final results.}
We use balanced accuracy, instead of edge accuracy because both \emph{Ensemble} and \emph{EnsembleHuman} are imbalanced, containing more \textit{negative} edges than \textit{positive} edges.
We plot the balanced accuracy results using both ensemble matrices (\emph{Ensemble} and \emph{EnsembleHuman}) as reference in Figure~\ref{benchmark_results_external}. 
We find that the two different evaluation results show a consistent ranking of baseline methods, although adding human annotations into evaluation slightly changes the absolute balanced accuracy for the baseline methods. Such a ranking consistency is a good evidence supporting our ensemble-based protocol for our clustering evaluation challenge.
% We find that even though human annotations will change the absolute balanced accuracy for the baseline methods within a certain range, it does not affect the ranking of these baseline methods. Therefore, we believe our Cluster3D evaluation protocol is suitable to be used to evaluate the ranking between different methods.

% We find that the human annotations have very slight influence on the final results, which is as expected because only about $6 \%$ of the human annotations, occupying $0.003 \%$ edges in Cluster3D, differ from and overwrite the \emph{Ensemble} matrix. A weighted version of this plot where human annotated edges are given higher evaluation weights can be found in the supplementary.
% \DP{I don't get why this is relevant, it is indeed a tiny difference. Is this just to justify why we have human annotations? It is very contrived and this will be pointed out in the reviews if we are not somehow defending this choice.}

\textit{End-to-end deep clustering methods do not necessarily outperform two-stage clustering methods.} As shown in Figure~\ref{benchmark_results_external}, for our task there is no major advantage for using end-to-end methods (DeepCluster, DEC, IIC, SCAN), compared to two-stage clustering methods (AtlasNet-based method, BYOL-based method, MVCNN-based method). Therefore, we believe it is necessary to study how to take the advantage of the clustering loss we use to train deep neural models.

\textit{The MVCNN-based method has the lowest accuracy.} We believe the different performance is due to the fact that it uses a different representation: images instead than point clouds. This suggests that in our dataset where class supervision does not exist and shape distribution is diverse, point-cloud-based unsupervised representation learning might be more effective than image-based ones.

\textbf{Clustering results using internal evaluation indices.}
Figure~\ref{benchmark_results_internal}-(Chamfer distance) shows the benchmark results using the \textit{silhouette score} as internal evaluation index. Using the Chamfer and Jaccard distances lead to similar  performances.
For all methods, we find that the clusters are not obviously different or even wrongly assigned, since most of the silhouette score is below 0. Another interesting observation is that the AtlasNet-based method, DEC, and SCAN methods perform better when the number of cluster $K$ increases, while other baseline methods show the opposite trend, which clearly shows which baselines perform better.
% Future investigation should be conducted to further understand this peculiar phenomenon.

\textbf{The baseline performances are mostly consistent in external and internal evaluation indices.} 
Comparing Figure~\ref{benchmark_results_external} and Figure~\ref{benchmark_results_internal}, we find that the ranking of the baseline methods are very consistent. The top three ranked methods are always AtlasNet-based method, SCAN, and DEC no matter using which indices. The DeepCluster and MVCNN-based methods are always performing badly in Cluster3D. This consistency is an additional evidence supporting our proposed ensemble-based evaluation protocol.
% \DP{check me}
% We believe with this ranking consistency, it is able to determine the method we want to use for deep clustering in Cluster3D.\DP{I do not understand this sentence.}

% \begin{figure}[htp]
%     \centering
%     \includegraphics[width=0.5\textwidth]{figs/tsne_placeholder.JPG}
%     \caption{\textbf{t-SNE}. This is an example figure that shows the t-SNE result. Please select cluster results for all baseline methods.}
%     \label{tsne_fig}
% \end{figure}

\subsection{Limitations and discussion}
The major challenge in our study is that it is not possible, for practical reasons, to have a fully annotated ground truth to rank the different baseline methods. Our approach based on an ensemble method is a proxy, justified by our experiments, that is likely to be a good approximation, but it is still an approximation. To increase the quality of this approximation, we can either add more data or more methods to improve the quality of the ensemble approximation, both directions that we are planning to explore as future work.

% The major challenge in our study is the very high cost of annotating a similarity matrix which has a quadratic number of entries with respect to the number of objects in the dataset. We introduced a technique to reduce the annotation cost, but it is possible that the filtering introduced a bias in the annotations. This bias could be reduced by picking multiple  initial clustering methods, which we plan to explore in the future. 

% The different evaluation metrics lead to different ranking for these baseline methods, suggesting that they evaluate different criteria. Identifying which metric is best for specific applications would be crucial to guide the development of clustering algorithms, and we believe it is an interesting venue for future work in deep clustering of 3D CAD models.

% DEC:
%  We use the standard AtlasNet architecture with an encoder based on PointNet. The encoder transforms the input point cloud of 4096(check) points into a latent vector of dimensions k=1024. The decoder consists of 4 fully-connected layers of size 1024, 512, 256 and 128 with ReLU non-linearities on the first three layer and tanh on the output layer. We train with output point clouds of 4096(check) size evenly sampled(how?check) across all the learned parametrications. To initalize centroids, we run k-means with 20 restarts and select the best solution. In the KL divergence minimization phase, we train with a constant learning rate of 0.001 and the convergence threshold is set to 0.1\%.
\section{Conclusion}\label{sec:conclusion}
This is the first work for the development and evaluation of clustering algorithms on collections of 3D CAD models. We introduce the annotated data, the ensemble matrix for evaluation, and evaluated seven clustering methods using both external and internal evaluation indices. We believe our work will be an important resource to develop and evaluate new clustering methods tailored for 3D geometry.

%  We introduce the dataset, two external evaluation metrics based on the matrix, and benchmarked seven state-of-the-art clustering methods. Our conclusion is that the gap between human annotators and state-of-the-art methods is large: we believe our dataset will be an important resource to improve clustering methods for 3D geometry.
% \input{parts/8-appendix}

{\small
\bibliographystyle{ieee_fullname}
\bibliography{egbib}

\begin{thebibliography}{10}\itemsep=-1pt

\bibitem[Bansal et~al.(2004)]{bansal2004correlation}
Nikhil Bansal, Avrim Blum, and Shuchi Chawla.
\newblock Correlation clustering.
\newblock {\em Machine learning}, 56(1):89--113, 2004.

\bibitem[Brodersen et~al.(2010)]{brodersen2010balanced}
Kay~Henning Brodersen, Cheng~Soon Ong, Klaas~Enno Stephan, and Joachim~M
  Buhmann.
\newblock The balanced accuracy and its posterior distribution.
\newblock In {\em 2010 20th international conference on pattern recognition},
  pages 3121--3124. IEEE, 2010.

\bibitem[Bustos et~al.(2007)]{4267943}
Benjamin Bustos, Daniel Keim, Dietmar Saupe, and Tobias Schreck.
\newblock Content-based 3d object retrieval.
\newblock {\em IEEE Computer Graphics and Applications}, 27(4):22--27, 2007.

\bibitem[Caron et~al.(2018)]{caron2018deep}
Mathilde Caron, Piotr Bojanowski, Armand Joulin, and Matthijs Douze.
\newblock Deep clustering for unsupervised learning of visual features.
\newblock In {\em Proceedings of the European Conference on Computer Vision
  (ECCV)}, pages 132--149, 2018.

\bibitem[Celebi(2014)]{celebi2014partitional}
M~Emre Celebi.
\newblock {\em Partitional clustering algorithms}.
\newblock Springer, 2014.

\bibitem[Chang et~al.(2015)]{chang2015shapenet}
Angel~X Chang, Thomas Funkhouser, Leonidas Guibas, Pat Hanrahan, Qixing Huang,
  Zimo Li, Silvio Savarese, Manolis Savva, Shuran Song, Hao Su, et~al.
\newblock Shapenet: An information-rich 3d model repository.
\newblock {\em arXiv preprint arXiv:1512.03012}, 2015.

\bibitem[Chen et~al.(2003)]{chen2003visual}
Ding-Yun Chen, Xiao-Pei Tian, Yu-Te Shen, and Ming Ouhyoung.
\newblock On visual similarity based 3d model retrieval.
\newblock In {\em Computer graphics forum}, volume~22, pages 223--232. Wiley
  Online Library, 2003.

\bibitem[Cutzu and Edelman(1998)]{cutzu1998representation}
Florin Cutzu and Shimon Edelman.
\newblock Representation of object similarity in human vision: psychophysics
  and a computational model.
\newblock {\em Vision research}, 38(15-16):2229--2257, 1998.

\bibitem[de~Beeck et~al.(2008)]{de2008perceived}
Hans P~Op de Beeck, Katrien Torfs, and Johan Wagemans.
\newblock Perceived shape similarity among unfamiliar objects and the
  organization of the human object vision pathway.
\newblock {\em Journal of Neuroscience}, 28(40):10111--10123, 2008.

\bibitem[Fan et~al.(2017)]{fan2017point}
Haoqiang Fan, Hao Su, and Leonidas~J Guibas.
\newblock A point set generation network for 3d object reconstruction from a
  single image.
\newblock In {\em Proceedings of the IEEE conference on computer vision and
  pattern recognition}, pages 605--613, 2017.

\bibitem[Funkhouser et~al.(2003)]{funkhouser2003search}
Thomas Funkhouser, Patrick Min, Michael Kazhdan, Joyce Chen, Alex Halderman,
  David Dobkin, and David Jacobs.
\newblock A search engine for 3d models.
\newblock {\em ACM Transactions on Graphics (TOG)}, 22(1):83--105, 2003.

\bibitem[Gao et~al.(2012)]{6200340}
Yue Gao, Meng Wang, Dacheng Tao, Rongrong Ji, and Qionghai Dai.
\newblock 3-d object retrieval and recognition with hypergraph analysis.
\newblock {\em IEEE Transactions on Image Processing}, 21(9):4290--4303, 2012.

\bibitem[Grill et~al.(2020)]{grill2020bootstrap}
Jean-Bastien Grill, Florian Strub, Florent Altch{\'e}, Corentin Tallec,
  Pierre~H Richemond, Elena Buchatskaya, Carl Doersch, Bernardo~Avila Pires,
  Zhaohan~Daniel Guo, Mohammad~Gheshlaghi Azar, et~al.
\newblock Bootstrap your own latent: A new approach to self-supervised
  learning.
\newblock {\em arXiv preprint arXiv:2006.07733}, 2020.

\bibitem[Groueix et~al.(2018)]{groueix2018papier}
Thibault Groueix, Matthew Fisher, Vladimir~G Kim, Bryan~C Russell, and Mathieu
  Aubry.
\newblock A papier-m{\^a}ch{\'e} approach to learning 3d surface generation.
\newblock In {\em Proceedings of the IEEE conference on computer vision and
  pattern recognition}, pages 216--224, 2018.

\bibitem[Hilaga et~al.(2001)]{hilaga2001topology}
Masaki Hilaga, Yoshihisa Shinagawa, Taku Kohmura, and Tosiyasu~L Kunii.
\newblock Topology matching for fully automatic similarity estimation of 3d
  shapes.
\newblock In {\em Proceedings of the 28th annual conference on Computer
  graphics and interactive techniques}, pages 203--212, 2001.

\bibitem[Huang and Liu(2019)]{10.1145/3343031.3351061}
Tianxin Huang and Yong Liu.
\newblock 3d point cloud geometry compression on deep learning.
\newblock Association for Computing Machinery, 2019.

\bibitem[Jatavallabhula et~al.(2019)]{Kaolin}
Krishna~Murthy Jatavallabhula, Edward Smith, Jean-Francois Lafleche,
  Clement~Fuji Tsang, Artem Rozantsev, Wenzheng Chen, Tommy Xiang, Rev
  Lebaredian, and Sanja Fidler.
\newblock Kaolin: A pytorch library for accelerating 3d deep learning research.
\newblock {\em arXiv:1911.05063}, 2019.

\bibitem[Ji et~al.(2019)]{ji2019invariant}
Xu Ji, Jo{\~a}o~F Henriques, and Andrea Vedaldi.
\newblock Invariant information clustering for unsupervised image
  classification and segmentation.
\newblock In {\em Proceedings of the IEEE/CVF International Conference on
  Computer Vision}, pages 9865--9874, 2019.

\bibitem[Koch et~al.(2019)]{koch2019abc}
Sebastian Koch, Albert Matveev, Zhongshi Jiang, Francis Williams, Alexey
  Artemov, Evgeny Burnaev, Marc Alexa, Denis Zorin, and Daniele Panozzo.
\newblock Abc: A big cad model dataset for geometric deep learning.
\newblock In {\em Proceedings of the IEEE/CVF Conference on Computer Vision and
  Pattern Recognition}, pages 9601--9611, 2019.

\bibitem[Kosub(2019)]{kosub2019note}
Sven Kosub.
\newblock A note on the triangle inequality for the jaccard distance.
\newblock {\em Pattern Recognition Letters}, 120:36--38, 2019.

\bibitem[MacQueen et~al.(1967)]{macqueen1967some}
James MacQueen et~al.
\newblock Some methods for classification and analysis of multivariate
  observations.
\newblock In {\em Proceedings of the fifth Berkeley symposium on mathematical
  statistics and probability}, volume~1, pages 281--297. Oakland, CA, USA,
  1967.

\bibitem[Osada et~al.(2002)]{osada2002shape}
Robert Osada, Thomas Funkhouser, Bernard Chazelle, and David Dobkin.
\newblock Shape distributions.
\newblock {\em ACM Transactions on Graphics (TOG)}, 21(4):807--832, 2002.

\bibitem[Paszke et~al.(2019)]{NEURIPS2019_9015}
Adam Paszke, Sam Gross, Francisco Massa, Adam Lerer, James Bradbury, Gregory
  Chanan, Trevor Killeen, Zeming Lin, Natalia Gimelshein, Luca Antiga, Alban
  Desmaison, Andreas Kopf, Edward Yang, Zachary DeVito, Martin Raison, Alykhan
  Tejani, Sasank Chilamkurthy, Benoit Steiner, Lu Fang, Junjie Bai, and Soumith
  Chintala.
\newblock Pytorch: An imperative style, high-performance deep learning library.
\newblock In H. Wallach, H. Larochelle, A. Beygelzimer, F. d\textquotesingle
  Alch\'{e}-Buc, E. Fox, and R. Garnett, editors, {\em Advances in Neural
  Information Processing Systems 32}, pages 8024--8035. Curran Associates,
  Inc., 2019.

\bibitem[Qi et~al.(2017)]{qi2017pointnet}
Charles~R Qi, Hao Su, Kaichun Mo, and Leonidas~J Guibas.
\newblock Pointnet: Deep learning on point sets for 3d classification and
  segmentation.
\newblock In {\em Proceedings of the IEEE conference on computer vision and
  pattern recognition}, pages 652--660, 2017.

\bibitem[Qi et~al.(2016)]{qi2016volumetric}
Charles~R Qi, Hao Su, Matthias Nie{\ss}ner, Angela Dai, Mengyuan Yan, and
  Leonidas~J Guibas.
\newblock Volumetric and multi-view cnns for object classification on 3d data.
\newblock In {\em Proceedings of the IEEE conference on computer vision and
  pattern recognition}, pages 5648--5656, 2016.

\bibitem[Rousseeuw(1987)]{rousseeuw1987silhouettes}
Peter~J Rousseeuw.
\newblock Silhouettes: a graphical aid to the interpretation and validation of
  cluster analysis.
\newblock {\em Journal of computational and applied mathematics}, 20:53--65,
  1987.

\bibitem[Shum et~al.(1996)]{shum19963d}
Heung-Yeung Shum, Martial Hebert, and Katsushi Ikeuchi.
\newblock On 3d shape similarity.
\newblock In {\em Proceedings CVPR IEEE Computer Society Conference on Computer
  Vision and Pattern Recognition}, pages 526--531. IEEE, 1996.

\bibitem[Socher et~al.(2012)]{socher2012convolutional}
Richard Socher, Brody Huval, Bharath Bath, Christopher~D Manning, and Andrew
  Ng.
\newblock Convolutional-recursive deep learning for 3d object classification.
\newblock {\em Advances in neural information processing systems}, 25, 2012.

\bibitem[Su et~al.(2015)]{su2015multi}
Hang Su, Subhransu Maji, Evangelos Kalogerakis, and Erik Learned-Miller.
\newblock Multi-view convolutional neural networks for 3d shape recognition.
\newblock In {\em Proceedings of the IEEE international conference on computer
  vision}, pages 945--953, 2015.

\bibitem[Sun et~al.(2019)]{8648155}
Xuebin Sun, Han Ma, Yuxiang Sun, and Ming Liu.
\newblock A novel point cloud compression algorithm based on clustering.
\newblock {\em IEEE Robotics and Automation Letters}, 4(2):2132--2139, 2019.

\bibitem[Van~Gansbeke et~al.(2020)]{van2020scan}
Wouter Van~Gansbeke, Simon Vandenhende, Stamatios Georgoulis, Marc Proesmans,
  and Luc Van~Gool.
\newblock Scan: Learning to classify images without labels.
\newblock In {\em European Conference on Computer Vision}, pages 268--285.
  Springer, 2020.

\bibitem[Wu et~al.(2015)]{wu20153d}
Zhirong Wu, Shuran Song, Aditya Khosla, Fisher Yu, Linguang Zhang, Xiaoou Tang,
  and Jianxiong Xiao.
\newblock 3d shapenets: A deep representation for volumetric shapes.
\newblock In {\em Proceedings of the IEEE conference on computer vision and
  pattern recognition}, pages 1912--1920, 2015.

\bibitem[Xie et~al.(2016)]{xie2016unsupervised}
Junyuan Xie, Ross Girshick, and Ali Farhadi.
\newblock Unsupervised deep embedding for clustering analysis.
\newblock In {\em International conference on machine learning}, pages
  478--487. PMLR, 2016.

\bibitem[Yang and Grzegorzek(2014)]{yang2014object}
Cong Yang and Marcin Grzegorzek.
\newblock Object similarity by humans and machines.
\newblock In {\em 2014 AAAI Fall Symposium Series}, 2014.

\bibitem[Zhan et~al.(2020)]{zhan2020online}
Xiaohang Zhan, Jiahao Xie, Ziwei Liu, Yew-Soon Ong, and Chen~Change Loy.
\newblock Online deep clustering for unsupervised representation learning.
\newblock In {\em Proceedings of the IEEE/CVF conference on computer vision and
  pattern recognition}, pages 6688--6697, 2020.

\bibitem[Zhang and Ma(2012)]{zhang2012ensemble}
Cha Zhang and Yunqian Ma.
\newblock {\em Ensemble machine learning: methods and applications}.
\newblock Springer, 2012.

\end{thebibliography}
}

\clearpage
%%%%%%%%%%%%%%%%%%%%%%%%%%%%%%%%%%%%%%%%%%%%%%%%%%%%%%%%%%%%
\def\ECCVSubNumber{4653}
% \title{Cluster3D: A Dataset and Protocol for\\Evaluating Clustering Algorithms for Non-Categorical 3D CAD Models
% }

\appendix

\section{Appendix}\label{sec:app}

\subsection{Challenges for class annotation}

The challenges for annotating CAD models in the ABC dataset are mainly from three aspects: (1) a large proConsistency between annotatorsrtion of non-standard models, (2) complex and unbalanced class distributions, and (3) lack of units and texture information. 

\begin{figure}[htp]
    \centering
    \includegraphics[width=0.4\textwidth]{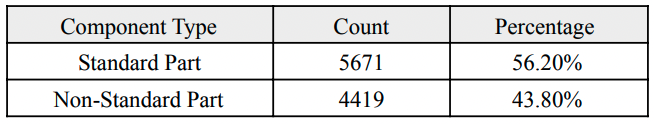}
    \caption{\textbf{Large proportion of non-standard models.}}
    \label{non-standard}
\end{figure}

\textbf{Large proConsistency between annotatorsrtion of non-standard models.}
We have four annotators to label $10,533$ CAD models from the ABC dataset. Each annotator will give the model a label, indicating it is a standard or non-standard model. We then collect all the label information and use a majority voting strategy to determine the final label for each model. Note that if the number of the standard label and non-standard label for a model is the same, we discard the model annotation. Finally, we have $10,090$ CAD models that can be considered to have a usable human label. The proConsistency between annotatorsrtion of the standard part and non-standard part can be seen in Table~\ref{non-standard}. We can see those non-standard models occupy around $45 \%$ of the data in our select dataset, showing that it is difficult to give class labels to almost half models in the dataset.

\begin{figure}[htp]
    \centering
    \includegraphics[width=0.4\textwidth]{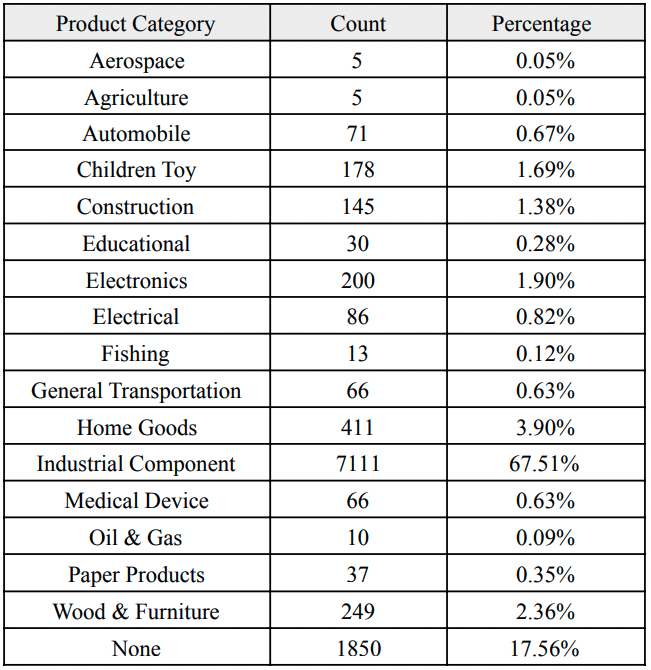}
    \caption{\textbf{Unbalanced class distributions.}}
    \label{unbalanced-class-distribution}
\end{figure}

\textbf{Complex and unbalanced class distributions.}
We also have four annotators to label the same $10,533$ CAD models for class. We can find that the category distribution in the dataset is unbalanced~\ref{unbalanced-class-distribution}. For example, industrial components occupy more than two-thirds of the models, while aerospace models and agriculture models only occupy $0.05 \%$. 

\textbf{Lack of units and texture information.}
During the labeling process, our annotators report that for some CAD models, it is difficult to give them labels because of the lack of units or texture information. 
% \begin{figure}[htp]
%     \centering
%     \includegraphics[width=0.45\textwidth]{figs/material.png}
%     \caption{\textbf{Lack of units and texture information.}}
%     \label{lack-texture-units}
% \end{figure}

\subsection{Interfaces for 3D models annotation}
\begin{figure*}[htp]
    \centering
    \includegraphics[width=1\textwidth]{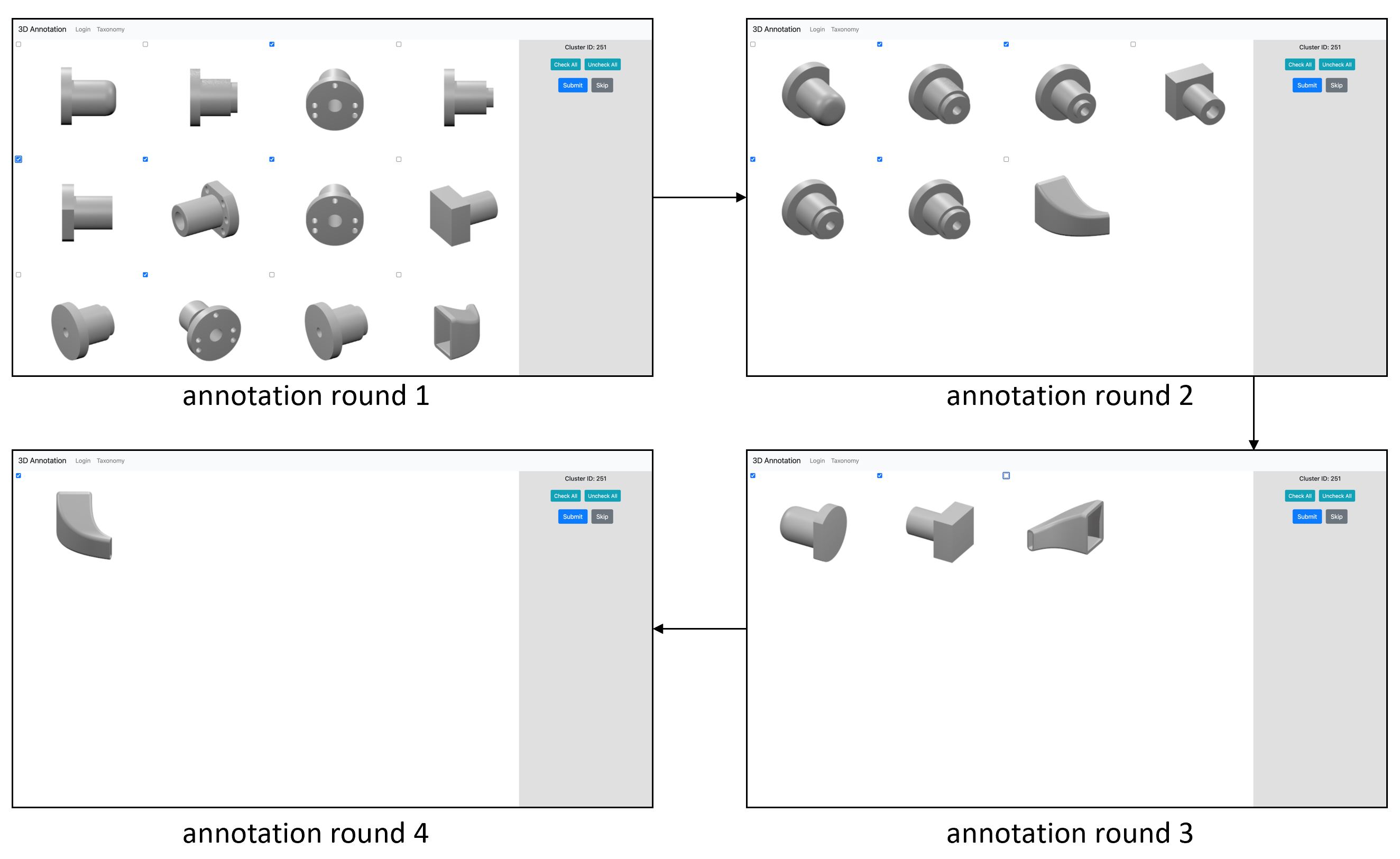}
    \caption{\textbf{Annotation interface example.}}
    \label{annotation_interface}
\end{figure*}

We show the example for annotating cluster 251 in Figure~\ref{annotation_interface}. For the annotation round 1, the interface shows 12 CAD models with checkboxes (set to be checked). The annotator should manually unmark the models that are not similar to the checked models. Then after round 1, the checked models will be labeled as similar to each other, and the remaining models will be sent to the annotation round 2. We repetitively ask the annotator to unmark the checks, until the annotator believes that the remaining models are all dissimilar to each other such that no sub-cluster can be created.

\subsection{Human annotator instruction and payment}
As aforementioned, the human standard to determine the similarity relationship between every two CAD models relies on previous expert knowledge that is difficult to describe. We provided the following instructions to our human annotators, to help them be more consistent. First, the similarity should be determined by the geometry, instead of the functionality. Second, the annotators should assume that the similarity relationship is rotational and translational invariant. %In another word, if the two CAD models are visually similar, only different in rotation or location, they should be considered as similar.
We pay the human annotators $10$ dollars per hour. In total, we spent around $12,000$ dollars for human annotations.

\subsection{Human annotation data statistics}
\begin{figure*}[htp]
    \centering
    \includegraphics[width=0.8\textwidth]{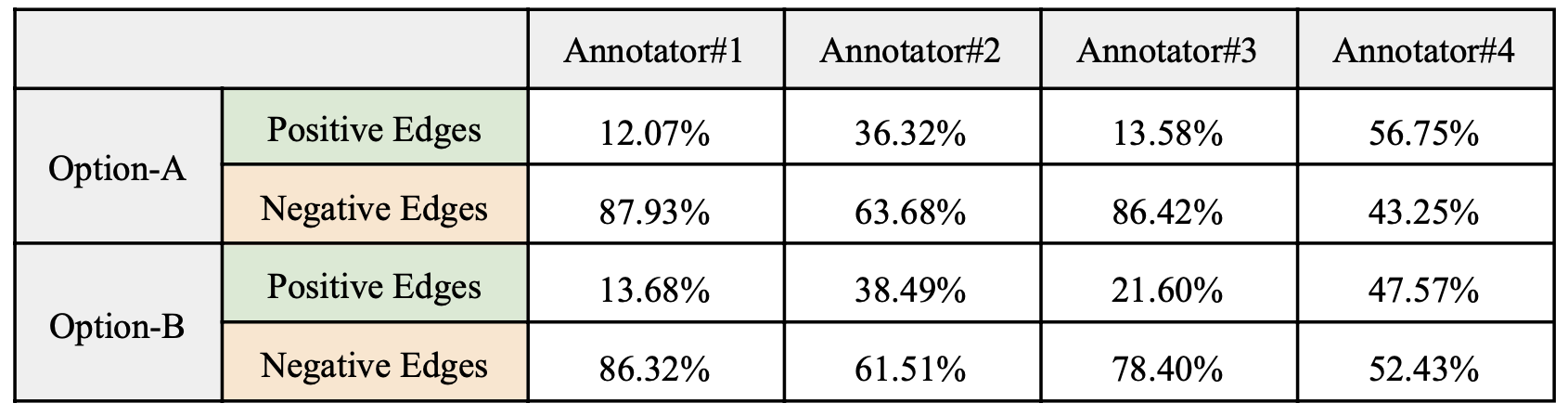}
    \caption{\textbf{Positive/negative edge percentage.}}
    \label{p_n_percentage}
\end{figure*}

\begin{figure*}[htp]
    \centering
    \includegraphics[width=0.8\textwidth]{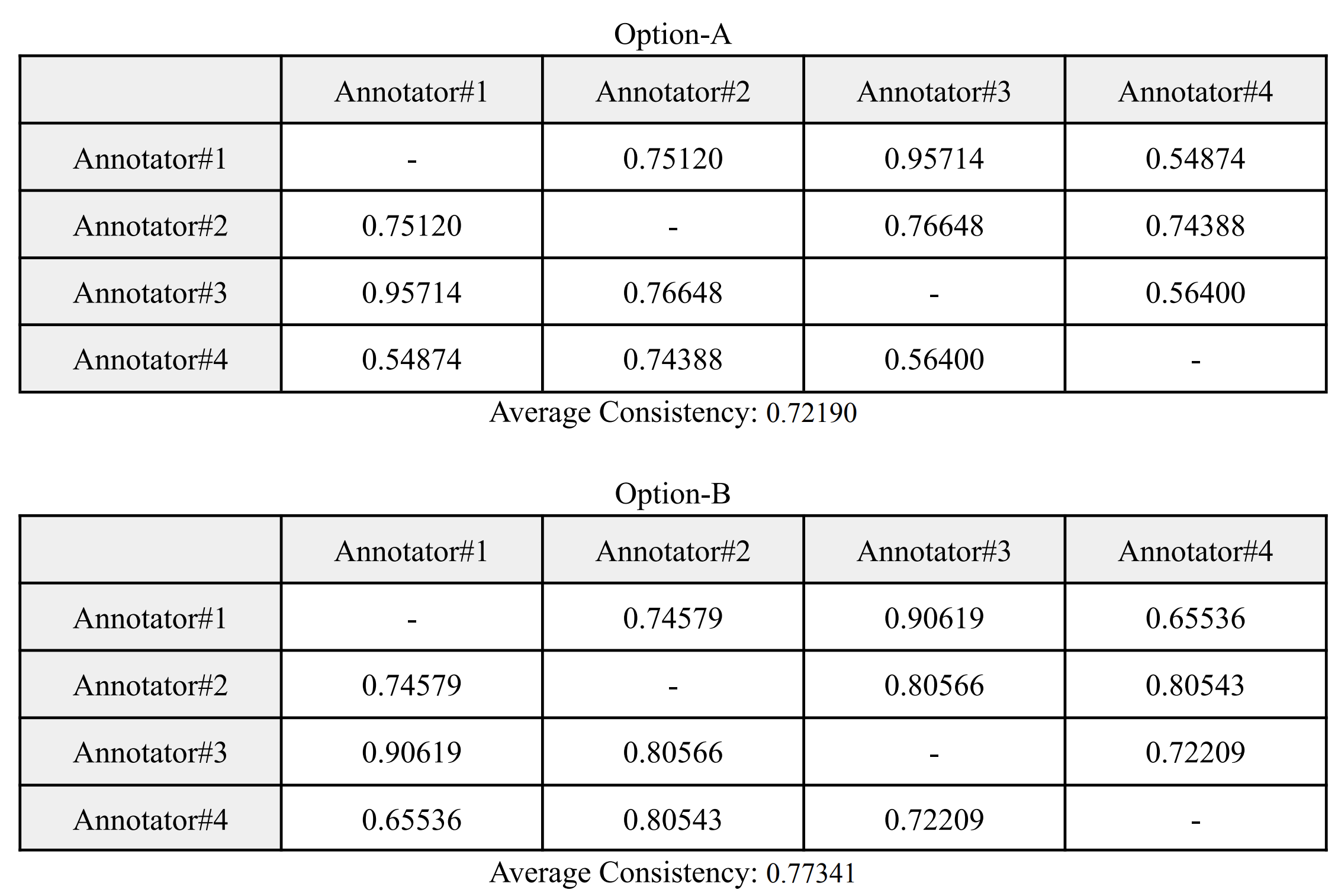}
    \caption{\textbf{Consistency between annotators.}}
    \label{label_consistency}
\end{figure*}

We show the Positive/nesitive/negative edge percentage of human annotation in Figure~\ref{p_n_percentage} and human label consistency in Figure~\ref{label_consistency} for both Option-A and Option-B. We find that: (1) different annotators might have different criteria to label the similarity relationships, since $Annotator \#1$ and $Annotator \#3$ give more negative labels than $Annotator \#2$ and $Annotator \#4$ (see Figure~\ref{p_n_percentage}); (2) Even though the annotators' annotation criteria could be different, most of the labeling consistency between annotators shows that human can achieve consensus on the 3D model similarity judgment (see Figure~\ref{label_consistency}). Therefore, we believe it is reasonable to use these human annotations.


\subsection{Data processing}

\textit{Point cloud generation.} We process 3D CAD models to point cloud by sampling $4,096$ points for each model, using Kaolin~\cite{Kaolin}. Then we apply min-max normalization to the sampled point cloud for each model.

\textit{Chamfer distance computation.}
The input of the Chamfer distance is the pre-processed point cloud for each CAD model that is normalized using min-max normalization (e.g. for each dimension, the min and max will be $0$ and $1$ after the normalization, and the scaling is affine).

\textit{Jaccard distance computation.}
Pairwise IoU (Intersection over Union, Jaccard index) of the objects are calculated first. The input of the IoU is voxel grids generated from the pre-processed point cloud for each CAD model that is normalized using min-max normalization. The IoU is not normalized for the case when the object is too thin to normalize, and thus the IoU of it versus anything will be zero. Then the Jaccard distance will be $1-\text{IoU}$ by definition.

\subsection{Network architectures}
For all the baseline methods, we use the original code from their reConsistency besitories and just modify the backbone networks to adjust to our data. We will release our code online and more details of the network could be found there.

\subsection{External evaluation indices}
We detail the \textit{edge accuracy} and \textit{balanced accuracy} in this part.

\begin{figure*}[htp]
    \centering
    \includegraphics[width=1\textwidth]{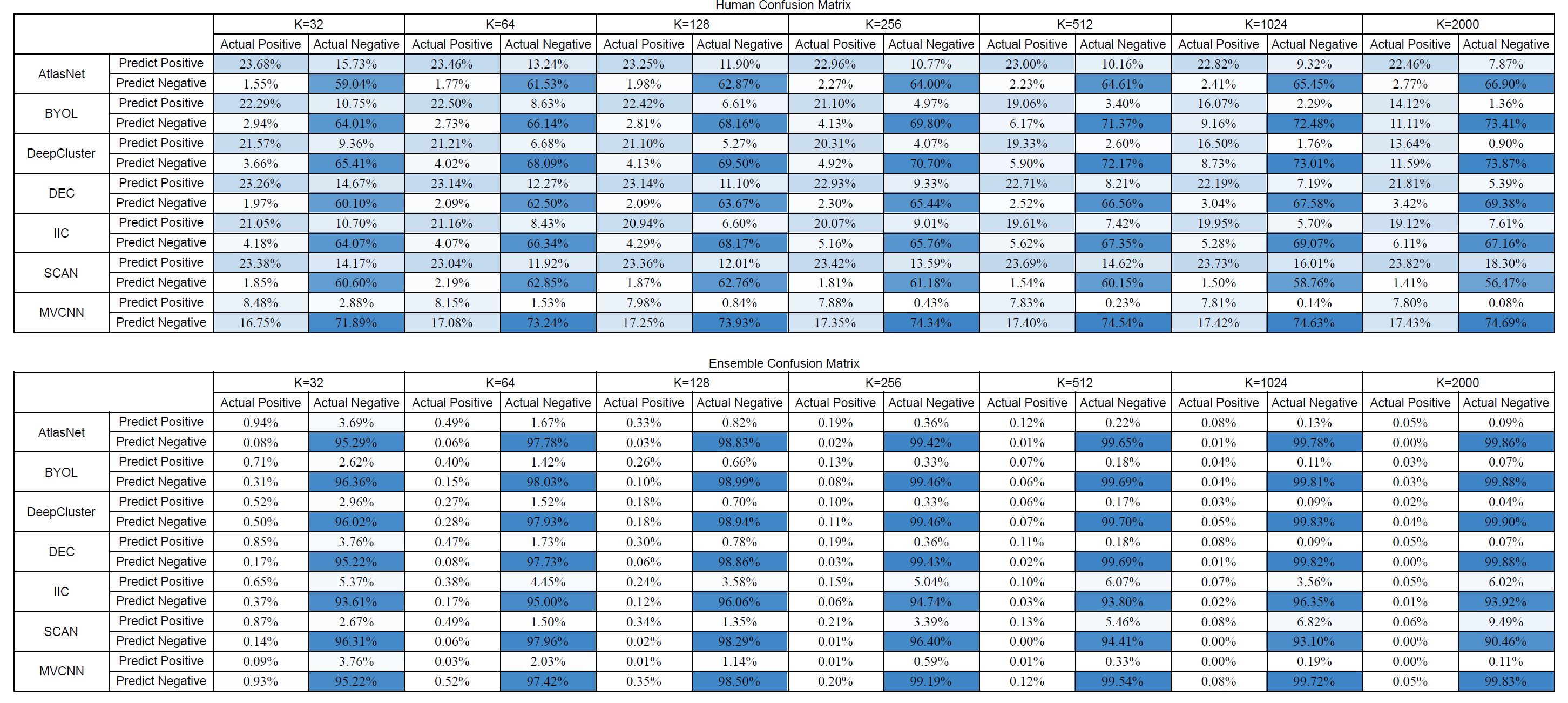}
    \caption{\textbf{Confusion matrices on \emph{Human} and \emph{Ensemble}.}}
    \label{confusion_matrix}
\end{figure*}

\textit{Edge accuracy.}
It is natural to compare the edge clustering results with the annotated similarity matrix, which is the evaluation metric in correlation clustering~\cite{bansal2004correlation}. We note that the similarity matrix is an undirected graph $\mathcal{G}(\mathcal{V}, \mathcal{E})$ on $N$ nodes. Let $e_{ij}$ denote the label of the edge relationship between object $i$, $j$, and $e_{ij} = e_{ji} $. $E = \left \{  e_{ij}\right \}$ denote all the edges.
$G^{\prime} = (V^{\prime}, E^{\prime})$ is the subgraph of of $G$, which is only comConsistency besed of the known labels. $E^{\prime} = \left \{  e_{ij}|e_{ij} = 1 \lor e_{ij}=-1, e_{ij}\in E \right \}$. %
For the clustering results obtained from the baseline methods, $\hat{e}_{ij}$ denote the clustered edge relationship between object $i$, $j$. If objects $i$, $j$ are grouped into the same cluster, we assume the two objects are similar, therefore $\hat{e}_{ij} = 1$. Otherwise $\hat{e}_{ij} = -1$.
The edge accuracy is defined as: acc = $\sum_{e_{ij}\in E^{\prime}}^{}\frac{\left | \hat{e}_{ij}-e_{ij} \right |}{2n(E^{\prime})}$, where $n(E^{\prime})$ is the number of elements in $E^{\prime}$. The range of the edge accuracy is $[0,1]$.

\textit{Balanced accuracy.}
Unlike \textit{edge accuracy} that might suffer from the unbalanced dataset, balanced accuracy can be used to better evaluate the methods' performance. Specifically, for each baseline method, after having all the edge relationships $\hat{e}_{ij}$ in $E^{\prime}$, we generate a confusion matrix that can be used to calculate balanced accuracy. The detailed information for balanced accuracy can be found in~\cite{brodersen2010balanced}.

\subsection{\emph{Human} and \emph{Ensemble} confusion matrices}

Table~\ref{confusion_matrix} shows all the confusion matrices generated using both \emph{Human} and \emph{Ensemble} matrices. Note that we use percentage representations in the confusion matrices for a more intuitive understanding of the performance of one baseline method.

\end{document}